\documentclass[letterpaper]{article} 
\usepackage{aaai2026}  
\usepackage{times}  
\usepackage{helvet}  
\usepackage{courier}  
\usepackage[hyphens]{url}  
\usepackage{graphicx} 
\usepackage{natbib}  
\usepackage{caption} 
\usepackage{algorithm}
\usepackage{algorithmic}
\usepackage{amsmath} 
\usepackage{booktabs} 
\usepackage{amssymb}
\usepackage{multirow}
\usepackage{newfloat}
\usepackage{listings}
\DeclareCaptionStyle{ruled}{labelfont=normalfont,labelsep=colon,strut=off} 
\floatstyle{ruled}
\newfloat{listing}{tb}{lst}{}
\floatname{listing}{Listing}
\nocopyright

\title{AutoLoRA: Automatic LoRA Retrieval and Fine-Grained Gated Fusion for Text-to-Image Generation}

\author {
    Zhiwen Li\textsuperscript{\rm 1},
    Zhongjie Duan\textsuperscript{\rm 2},
    Die Chen\textsuperscript{\rm 1},
    Cen Chen\textsuperscript{\rm 1},
    Daoyuan Chen\textsuperscript{\rm 2},
    Yaliang Li\textsuperscript{\rm 2},
    Yingda Chen\textsuperscript{\rm 2}
}
\affiliations {
    \textsuperscript{\rm 1}East China Normal University\\
    \textsuperscript{\rm 2}Alibaba Group\\
    \{zhiwenli, dchen\}@stu.ecnu.edu.cn, cenchen@dase.ecnu.edu.cn, \\
    \{duanzhongjie.dzj, daoyuanchen.cdy, yaliang.li, yingda.chen\}@alibaba-inc.com
}

\usepackage{bibentry}

\begin{document}

\maketitle

\begin{abstract}
Despite recent advances in photorealistic image generation through large-scale models like FLUX and Stable Diffusion v3, the practical deployment of these architectures remains constrained by their inherent intractability to parameter fine-tuning. While low-rank adaptation (LoRA) have demonstrated efficacy in enabling model customization with minimal parameter overhead, the effective utilization of distributed open-source LoRA modules faces three critical challenges: sparse metadata annotation, the requirement for zero-shot adaptation capabilities, and suboptimal fusion strategies for multi-LoRA fusion strategies. To address these limitations, we introduce a novel framework that enables semantic-driven LoRA retrieval and dynamic aggregation through two key components: (1) weight encoding-base LoRA retriever that establishes a shared semantic space between LoRA parameter matrices and text prompts, eliminating dependence on original training data, and (2) fine-grained gated fusion mechanism that computes context-specific fusion weights across network layers and diffusion timesteps to optimally integrate multiple LoRA modules during generation. Our approach achieves significant improvement in image generation perfermance, thereby facilitating scalable and data-efficient enhancement of foundational models. This work establishes a critical bridge between the fragmented landscape of community-developed LoRAs and practical deployment requirements, enabling collaborative model evolution through standardized adapter integration.
\end{abstract}

\section{Itroduction}

The rapid advancement of foundation models for image generation has yielded astonishing capabilities in producing photorealistic and diverse imagery \cite{Robin2022ldm,Patrik204sd3,bf2024flux}. However, this progress has been accompanied by an exponential proliferation of model parameters, making fine-tuning increasingly challenging. To address this, the community widely adopts LoRA \cite{Edward2022lora}—a cost-effective and efficient fine-tuning technique—for personalized model customization. Many developers actively share their trained LoRA models on open-source platforms (such as ModelScope, Hugging Face, and Civitai), significantly accelerating community growth and catalyzing innovative AI-generated art.Today, thousands of LoRA models thrive in open-source repositories, with numbers continuously expanding. These adapters essentially reveal gaps in foundation models' capabilities, If these LoRAs can be fully utilized, we can further improve the performance of the basic model.Yet harnessing this vast LoRA ecosystem presents a formidable challenge. An optimal solution involves implementing a dynamic LoRA retrieval system that automatically selects contextually relevant adapters based on text prompts during image generation, coupled with a seamless LoRA fusion mechanism to integrate multiple adapters. This approach maximizes the utility of existing adapters to enhance base model performance while ensuring developers' contributions are discoverable and impactful. Consequently, it fosters community engagement, incentivizes the creation of higher-quality LoRA models, and establishes a self-reinforcing virtuous cycle of innovation.

Our goal is to build a framework for LoRA retrieval and fusion that can fully leverage existing LoRAs in the open source community and enhance various aspects of the text-to-image model. However, retrieving and fusing relevant LoRAs presents unique challenges compared to other RAG systems \cite{Patrick2020ragnlp,Yunfan2023ragsurvey}.
Specifically, in order to effectively perform LoRA retrieval, LoRA needs to be represented in an embedding form.  While PHATGOOSE \cite{Mohammed2024PHATGOOSE} is training the LoRA model, an embedding vector is also trained as the representation of the LoRA model. SemLA \cite{Reza2025SemLA} explores using the image embeddings in the LoRA training set to represent LoRA in the image segmentation system.
Unfortunately, we often cannot access the training data set of the LoRAs on the open source platform, and there is usually a lack of detailed documentation about the LoRAs, which makes it difficult for us to use traditional document-based retrieval enhancement solutions. Secondly, in this scenario, the pool of LoRAs is incrementally updated, and zero-shot adaptation for newly added LoRAs is crucial for the retrieval system.
In addition, the compatibility of multiple LoRAs is also a major challenge. Simple linear combinations or distilling multiple LoRAs into one model will suppress the performance of a single LoRA. As the number of aggregated adapters increases, the effect is difficult to guarantee. Some studies \cite{Xun2025MoLE, Ziyu2024RAMoLE} have explored the use of the MOE mechanism to integrate multiple LoRAs, but these methods limit the number of LoRA fusions and cannot adapt to any different LoRA aggregation situations. And according to recent research \cite{Ziheng2025Klora}, different LoRAs in the diffusion model will play a role in different diffusion steps and different model layers, which also poses a challenge to effective fusion.

To address the above challenges, we propose \textit{AutoLoRA}, a LoRA retrieval and dynamic aggregation framework, which can leverage the capabilities of the LoRA models on open source platforms to improve the performance of text-to-image models. This framework consists of two key components:
(1) \textbf{Weight encoding-base LoRA retriever}: Given that the CLIP model structure has achieved great success in the field of multimodal representation and retrieval, we created a LoRA weight encoder that encodes LoRA weight information and text information into the same feature space through contrastive learning. Since the model weights are different from those of general text images, it is difficult to extract features using conventional methods. Therefore, we regard the weight matrix of each layer of LoRA as a token, convert each token into a token embedding through some trainable parameters, and finally extract features through the transformer layer to obtain the embedding representation of the entire LoRA.
(2) \textbf{Fine-grained dynamic gated fusion mechanism: }After retrieving the relevant LoRAs, we need to carefully combine them to maximize their effect when generating images. In order to overcome the limitations of the traditional combination methods mentioned above, we propose a novel fine-grained dynamic gating LoRA fusion mechanism based on the generation characteristics of the diffusion model. In each linear layer of the model, different learnable gating modules are used to perceive the hidden state features of each intermediate layer of the original model and the LoRA model during the diffusion process, and the weights of each LoRA in different dimensions are calculated in real time. Unlike the naive LoRA combination method that assigns the same fixed weight to each LoRA, our fusion mechanism flexibly and subtly adjusts the LoRA weights in different feature dimensions at each intermediate layer to enhance the overall performance.

In order to evaluate the utility of the proposed framework, we downloaded 162 popular FLUX models of LoRA from different open source platforms. These LoRAs cover various themes and tasks, and they have different structures. Although we cannot obtain the training data and detailed description documents of these models, each LoRA model on the open source platform usually comes with several renderings. We use the VL model to generate corresponding text for each rendering to build a training set and test set. Experiments show that only a small number of LoRA renderings are needed to train our LoRA encoder and gated fusion module, and the quality of the generated images can be improved by combining the retrieved LoRAs.

In summary, our main contributions are:
\begin{itemize}
    \item We propose a retrieval model based on LoRA weight encoding, which encodes the weight parameters of LoRA as an embedding layer, maps LoRA representation and text representation to the same space through contrastive learning, and realizes semantic-driven LoRA model retrieval, which is expected to take advantage of the growing ecological model on the open source platform.
    \item We propose a novel fine-grained dynamic gated fusion mechanism, which realizes the flexible aggregation of multiple LoRA models through learnable gating modules, and improves the quality and stability of any number of LoRA collaborative generation.
    \item Empirical results demonstrate that the proposed AutoLoRA framework improves the overall performance of base text-to-image models by retrieving relevant LoRA models and efficiently aggregating them.
\end{itemize}

\section{Related Works}
\subsection{Adapter Retrieval}
As Adapter technology matures, the proliferation of Adapters within the community has prompted growing research interest in retrieving task-specific Adapters. A straightforward approach involves leveraging traditional Mixture of Experts (MoE) \cite{LepikhinLXCFHKS2021GSHard,FedusZS2022SwitchTransformers}techniques for Adapter retrieval, where methods like SMEAR \cite{MuqeethLR24SMEAR} and MoLE\cite{Xun2025MoLE} introduce an additional routing module during training. This module dynamically directs input tokens to different adapter experts based on their semantic content.The primary limitation of MoE-based approaches is that the number of experts remains fixed and typically limited, resulting in insufficient flexibility and scalability. SemLA \cite{Reza2025SemLA} proposes a training-free framework for image segmentation that retrieves adapters by measuring similarity between the input image and the adapter training dataset. PHATGOOSE \cite{Mohammed2024PHATGOOSE}, in contrast, trains a LoRA signature vector using the training dataset during LoRA model training and subsequently retrieves adapters by computing the similarity between input tokens and the signature vector, thereby enhancing LLM performance across diverse tasks and improving adaptability in zero-shot scenarios. RAMoLE \cite{Ziyu2024RAMoLE} trains a sentence-embedding model through instruction fine-tuning and retrieves task-specific LoRA models in LLMs by analyzing the similarity of sentence embeddings derived from input text. Although these methods demonstrate effectiveness to some extent, most are specifically designed for LLMs, making them challenging to adapt to image generation domains, and they universally require access to adapters' original training data—significantly restricting their generalizability.

\subsection{Adapter Fusion}
In the field of image generation, integrating personalized Adapters has long been a significant challenge, prompting numerous studies to be proposed \cite{Jiahua2024flexible,Yang2024loracomposer,Yuchao2024mixofshow}. ZipLoRA \cite{Viraj2024ziplora} proposes a straightforward method to learn scaling coefficients that render the columns of two adapters' weight matrices nearly orthogonal, thereby preventing interference when combining the adapters. Inspired by recent advances in Mixture of Experts (MoE) techniques, MoLE \cite{Ziyu2024RAMoLE} treats each adapter as an expert module and trains a gating mechanism within every feed-forward network layer to dynamically modulate the contribution of each adapter across different model layers. Unlike methods that combine different models through integration, LoRACLR \cite{Enis2025loraclr} distills knowledge from multiple target-concept adapter models across various image generation frameworks into a single adapter, enabling accurate generation of images depicting multiple concepts simultaneously. the recent K-LoRA \cite{Ziheng2025Klora}  propose a training-free approach that calculates the top-k elements from each target model within every attention layer to dynamically determine which model should be activated at each step. Similarly, DARE \cite{Leyu2024dare} employs a training-free strategy inspired by stochastic dropout, randomly discarding incremental parameters from different models according to specific policies to mitigate conflicts during model fusion.
Although effective in controlled settings, these methods typically focus on fusing only a small number of fixed LoRAs. They still encounter critical limitations—including poor subject consistency and high training complexity—making them impractical for scenarios where we dynamically retrieve arbitrary numbers of diverse LoRAs for each prompt.

\begin{figure*}
    \centering
    \includegraphics[width=1.\linewidth]{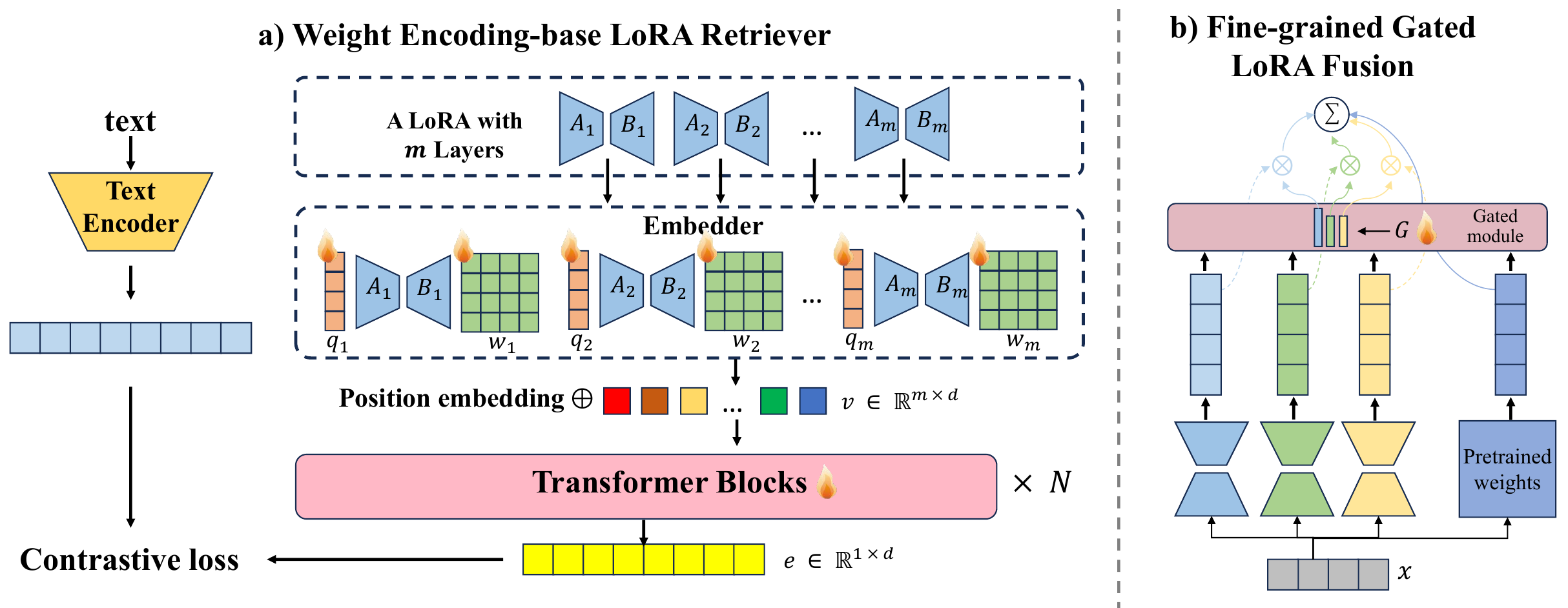}
    \caption{Illustration of the AutoLoRA Framework.
    It comprises two key components:
a) \textbf{Weight Encoding-Based LoRA Retriever:} Leverages a LoRA weight encoder to map both LoRA weights and textual prompts into a shared feature space through contrastive learning.
b) \textbf{Fine-Grained Gated LoRA Fusion:} Employs learnable gated modules to dynamically compute LoRA weights across different dimensions, thereby optimizing the collaborative performance of multiple LoRAs during fusion.}
    \label{fig:method}
\end{figure*}

\section{AutoLoRA Framework}
In this section, we present a detailed overview of the proposed AutoLoRA Framework, as illustrated in Figure \ref{fig:method}. It consists of two major components: Weight Encoding-based LoRA Retriever and Fine-grained gated fusion mechanism.
\subsection{Weight Encoding-base LoRA Retriever}
Our goal is that the user inputs a text prompt, and the retriever can recall k LoRAs associated with the text prompt from the lora pool, and the number of LoRAs in the pool of LoRAs is incrementally updated. To achieve this goal, we use a CLIP \cite{Alec2021clip} model architecture which contain a text encoder and a LoRA encoder, they can encode text and LoRA into an embedding respectively, and then calculate the similarity between the embeddings to complete the retrieval. The text encoder we can use pre-trained model, but how do we design a LoRA encoder that can input a LoRA weight parameter and output an embedding?

As we all know, LoRA introduce low-rank matrices to each linear layer. Specifically, for a linear with weight $\mathbf{W_0} \in \mathbb{R}^{d \times k}$, LoRA augment it with two low-rank matrics: $W_{0}+\Delta W=W_{0}+B A$, where $B \in \mathbb{R}^{d \times r}, A \in \mathbb{R}^{r \times k}$and $r \ll \min (d, k)$. Further, we denote a LoRA as $\Delta W = (B, A)$, where $B = \{B_1, \cdots, B_m  \}$ and $A = \{ A_1, \cdots, A_m \}$ denote LoRA is applied to $m$ linear layers of the original model. We regard each individual LoRA layer $(B_i, A_i)$ as a token and first convert it into a token embedding:
\begin{equation}
    v_i = B_i A_i q_i \hat{W}_i .
\end{equation}
where $q_i \in \mathbb{R}^{k}$ is a trainable parameter and $\hat{W}_i \in \mathbb{R}^{d \times out\_dim}$ is a learnable matrix used to convert each individual LoRA layer embedding into the same dimension.
After encoding all LoRA layers, we can obtain an embedding sequence $v \in \mathbb{R}^{m \times out\_dim}$, and this process can be denoted as: $v = \text{Embeder((B,A))}$. In order to obtain the global features of LoRA, we use Transformer Blocks to further extract features:
\begin{equation}
    e = \text{LoRAEncoder}(v).
\end{equation}
where $e \in \mathbb{R}^{1 \times out\_dim}$ is the final representation embedding of LoRA and $\text{Encoder}(\cdot)$ represents the encoder composed of $n$ standard Transform Blocks.
The overall model structure of LoRA Encoder is shown in the Fig \ref{fig:method}. 

\subsubsection{Training Object}We use contrastive learning for training. First, use VLM to convert each LoRA rendering into text, and use these texts as the labels corresponding to each LoRA. In this way, we can get a data set $\mathcal{T} = \{(x_{1,1},(B,A)_1),  \cdots, (x_{j,t},(B,A)_j)\}$, where each LoRA corresponds to at least one text description.
We use the pre-trained CLIP text encoder to encode text: $t = \text{CLIP}_{x}(text)$, and freeze the parameters of the text encoder during training, only update the parameters of the LoRA encoder, and take $N$ data from $\mathcal{T}$ for training each iteration. The loss function is as follows:
\begin{equation}
    \mathcal{L} = \sum_{i=1}^N \left( 
-\log \frac{\exp(e_i^\top t_i)}{\sum_{j=1}^N \exp(e_i^\top t_j)} 
- \log \frac{\exp(e_i^\top t_i)}{\sum_{j=1}^N \exp(e_j^\top t_i)} 
\right)
\end{equation}
In the LoRA retrieval stage, given a LoRA set $\Phi$ and an input prompt $x$, the $top-k$ LoRAs are retrieved according to the cosine similarity:
\begin{equation}
    s = cos(\text{LoRAEncoder((B,A))}, \text{CLIP}_{text}(x)).
\end{equation}
This process can be expressed as:
\begin{equation}
    \Phi_k = \text{TopK}\{ s((B,A)_j,x), (B,A)_j \in \Phi \}
\end{equation}

\subsection{Fine-grained Dynamic Gated LoRA Fusion}
After retrieving $k$ relevant LoRAs, the next thing we need to do is to integrate these LoRAs into the diffusion model. A straightforward idea is to use the MoE-base method \cite{Xun2025MoLE} to train an additional router in each layer and assign different weights to each LoRA. However, the traditional MoE method requires a fixed number of LoRAs during training, and can only assign weights to these fixed LoRAs during inference. It cannot be applied to the scenario of dynamically selecting LoRAs from the LoRA pool for fusion. To address this challenge, we propose a Fine-grained dynamic gating LoRA fusion mechanism, which utilizes a learnable gating module in the linear layer to perceive the hidden state features of each intermediate layer of the original model and LoRA moel during the diffusion process, and dynamically calculates the LoRA model weights of different dimensions.

Formally, consider top-k retrieved LoRAs in a linear layer. The output of original model output $\mathbf{x} \in\mathbb{R}^{l \times d}$ and a collection of LoRA outputs $\mathbf{L} = [\mathbf{l}_1, \mathbf{l}_2, \dots, \mathbf{l}_k] \in\mathbb{R}^{k \times l \times d}$  , the module first applies LayerNorm normalization to both inputs to eliminate scale discrepancies and highlight critical features: 
\begin{equation}
\hat{\mathbf{x}} = \text{LayerNorm}(\mathbf{x}), \quad\hat{\mathbf{L}} = \text{LayerNorm}(\mathbf{L}) 
\end{equation}
Subsequently, our gating mechanism computes dimension-specific contribution weights for each LoRA through the synergistic operation of three specialized gate components:
\begin{equation}\mathbf{G} = \sigma\left( \hat{\mathbf{x}} \odot\mathbf{w}_x + \hat{\mathbf{L}} \odot\mathbf{w}_l + \hat{\mathbf{x}} \odot\hat{\mathbf{L}} \odot\mathbf{w}_c + \mathbf{b} \right). \end{equation}
where $\mathbf{w}_x$ represents the \textbf{Base Feature Gate} which models the dominant representation importance of the original output, $\mathbf{w}_l$ denotes the \textbf{LoRA-Specific Gate} which senses the differentiated contribution of each LoRA adapter, $\mathbf{w}_c$ constitutes the \textbf{Cross-Interaction Gate} which captures the high-order interaction features between the original output and the LoRA output, all three are learnable weight vectors $\mathbb{R}^d$. And $\mathbf{b} \in \mathbb{R}^d$ is a learnable bias term, $\sigma$ denotes the sigmoid activation function, and $\odot$ represents element-wise multiplication.
The resulting gating matrix $\mathbf{G} \in \mathbb{R}^{k \times d}$  contains dynamic weights $g_{i,d}$ that determine the contribution of the $i$-th LoRA at dimension $d$, with values determined through the collaborative decision-making of the three gate components. Finally, the module integrates the original output with the weighted LoRA outputs through amplitude calibration:
\begin{equation}
\mathbf{x'} = \mathbf{x} + \sum_{i=1}^{k} \left( \mathbf{w}_o \odot\mathbf{g}_i \odot\mathbf{l}_i \right) 
\end{equation}

where $\mathbf{w}_o$ is a learnable \textbf{Fusion-Scaling Parameter} which further ensures numerical stability during integration and $\mathbf{g}_i$ represents the $i$-th row of $\mathbf{G}$.

\subsubsection{Global LoRA} Inspired by some MoE approaches \cite{Damai2024deepseekmoe} who shared experts capture and integrate general knowledge across diverse contexts—we introduce a global general LoRA during LoRA fusion. This LoRA is not a standalone model but is synthesized by summing the weight matrices of the target LoRAs and decomposing the resulting matrix into two low-rank components via matrix decomposition. The constructed global LoRA encapsulates cross-contextual information from the fused models and is seamlessly integrated into both training and generation phases:
\begin{equation}
    A_g B_g = \mathcal{D}_{r_g} \left( \sum_{i=1}^{n} B_i A_i \right).
\end{equation}

where  $\mathcal{D}_{r_g}(\cdot)$ denotes the matrix decomposition algorithm; we employ SVD for matrix decomposition, and to improve computational efficiency, we leverage PyTorch's approximate SVD implementation by setting the rank of the decomposed matrix to a small value (typically 4). This significantly reduces computational costs while ensuring efficiency in both training and inference phases

\subsubsection{Training Object} We use a strategy called Interference-Resistant Training to train this fusion module. During each training iteration, we randomly sample two LoRA adapters from the pool: one designated as the \textbf{target LoRA} $L_i$ and the other as the \textbf{interference LoRA} $L_j$. While both LoRAs are simultaneously active in the forward pass, the training signal is exclusively derived from the target LoRA's image-text pairs.This asymmetric supervision forces the fusion module to learn discriminative gating behavior---effectively amplifying relevant features while suppressing interference---despite the concurrent presence of both adapters. The optimization objective employs loss function of flow matching: 
\begin{equation}
    L=\mathbb{E}\left[\left\|V_{\hat{\theta}}\left(x_t,c,t,L_i,L_j,L_g\right)-\left(x_1-x_0\right)\right\|^2\right]
\end{equation}

where $V_{\hat{\theta}}$ is the diffusion model with LoRA fusion mechanism, $L_g$ is a global general LoRA constructed by $L_i$and $L_j$,  $x_t$is the targer image in laten space, $x_0 \sim N(0,1)$is the noise, $c$is the text condition, and $t \sim \mu(0,1)$is the timestep.

\begin{figure*}[t]
    \centering
    \includegraphics[width=0.99\linewidth]{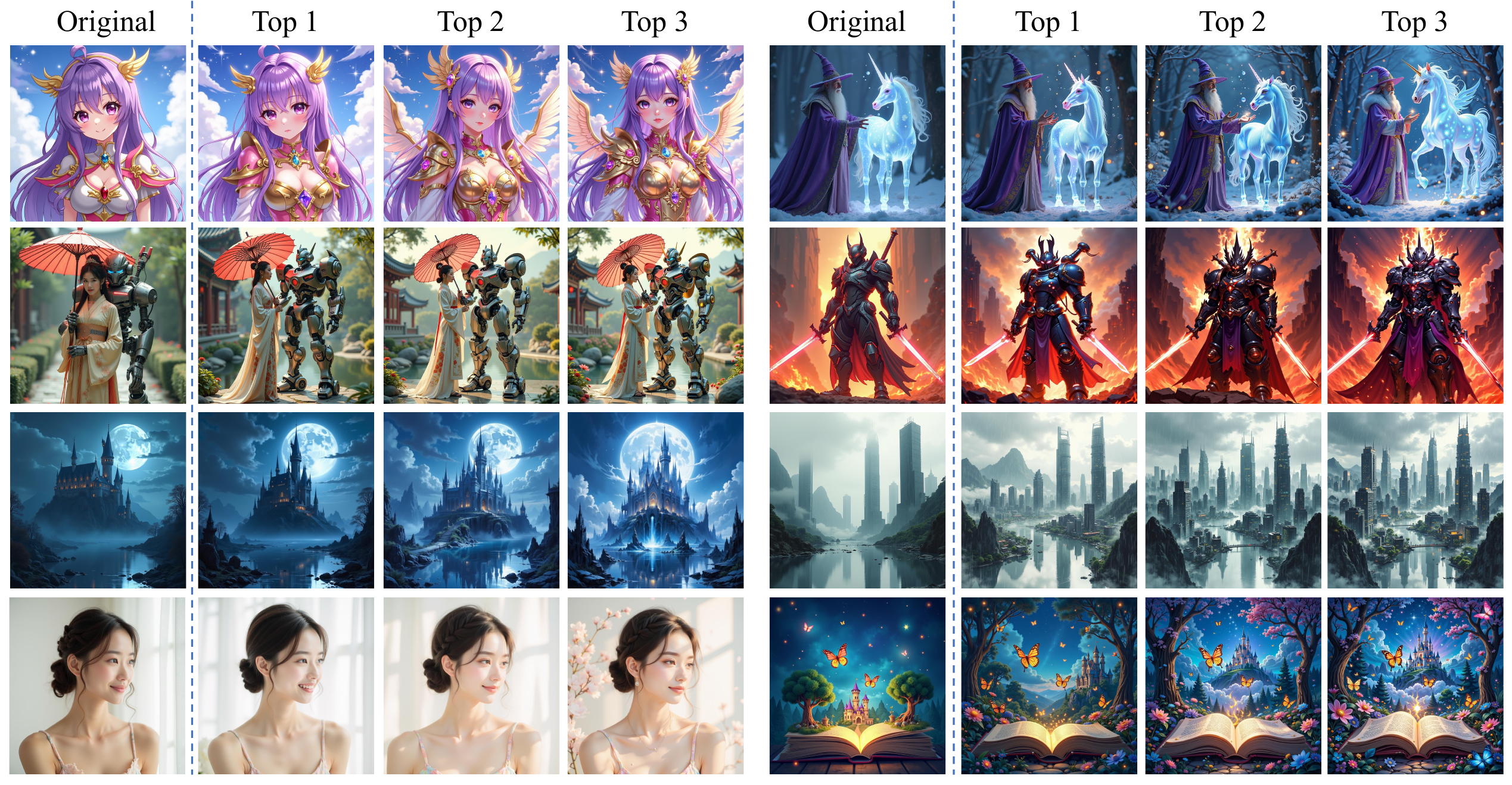}
    \caption{Qualitative comparison between AutoLoRA and original FLUX.1-dev. Top-1, Top-2, and Top-3 represent images generated after retrieving and fusing 1, 2, and 3 LoRAs, respectively. AutoLoRA can enriches visual details, refines artistic characteristics, and boosts the image's overall aesthetics}
    \label{fig:retrival-fusion}
\end{figure*}

\section{Experiments}
\subsection{Retrieval-Augmented Fusion for Image Generation}
In this section, we conduct a comprehensive evaluation of the AutoLoRA framework. 
Given an input prompt, we first use the LoRA Retriever to search for $k$ related LoRAs from the pool and then apply our gated fusion mechanism during generation to integrate their capabilities.
\subsubsection{Experimental Setup}
To comprehensively demonstrate AutoLoRA's capabilities, we downloaded 162 popular FLUX.1-dev LoRA models of diverse types from multiple open-source platforms, thereby forming our LoRA pool. This dataset exclusively includes the LoRA weight parameters and a small number of images (1–5 per LoRA) generated by the respective LoRA. We then utilized Qwen VL to generate descriptive captions for each image. For the LoRA retrieval experiments, we randomly selected 1–3 images from the dataset and employed Qwen VL to create prompts that simultaneously capture features from multiple images, ultimately constructing a synthetic prompt set comprising 900 prompts in total. To evaluate the framework's generalizability, we also randomly sampled 1,000 prompts from DiffusionDB \cite{Wang2023diffusiondb} and rewrote them using a LLM to construct a dedicated dataset. The evaluation metrics employed encompass image aesthetic score: MPS \cite{Zhang2024MPS}, HPS \cite{Xiaoshi2023HPS}, Aesthetic \cite{Schuhmann2022aes} and alongside the text alignment score: VQAScore \cite{Zhiqiu2024vqascore}. For each prompt, we retrieved the top-1, top-2, and top-3 LoRAs to generate images.

\subsubsection{Result}
Table \ref{tab:retrieve-and-fusion} presents the quantitative results of AutoLoRA on the two datasets. As evident from the table, AutoLoRA enhances the aesthetic quality of generated images compared to the original FLUX model, demonstrating that suitable LoRAs for the input prompt have been successfully retrieved and their performance maximized through the fusion mechanism—thereby compensating for the original FLUX model's lack of corresponding capabilities.Concurrently retrieved LoRAs also enhance text alignment, as these relevant adapters help the model better interpret and focus on specific elements within the prompt, thereby generating images that more strictly adhere to textual instructions. Our observations further reveal that while performance improves with an increasing number of retrieved LoRAs, diminishing marginal returns become increasingly evident. Additional LoRAs yield progressively smaller gains—primarily because retrieval is similarity-based: the top-1 LoRA exhibits the highest relevance to the prompt and delivers the most significant improvement, while subsequent LoRAs (with lower similarity scores) contribute diminishing benefits. The results on the DiffusionDB dataset demonstrate that AutoLoRA remains effective even for out-of-distribution prompts, highlighting the strong generalizability of our proposed framework. Figure \ref{fig:retrival-fusion} shows some visualization results, compared to the baseline FLUX.1-dev outputs, integrating retrieved LoRAs significantly demonstrates enhances visual fidelity by enriching structural details, elevating artistic style coherence, and improving overall aesthetic quality in generated images

\begin{table}[ht]
\centering
\small
\caption{Quantitative experimental results using AutoLoRA on synthetic prompt set and DiffusionDB}
\label{tab:retrieve-and-fusion}
\setlength{\tabcolsep}{2.5pt}
\begin{tabular}{c|c|c|cccc} 

\toprule
Dataset                                                                         & \multicolumn{2}{c|}{Method}       & MPS    & HPS    & Aes.  & VQA   \\ 
\midrule
\multirow{4}{*}{\begin{tabular}[c]{@{}c@{}}Synthetic \\prompt set\end{tabular}} & \multicolumn{2}{c|}{FLUX.1-dev} & 17.294 & 0.324 & 6.302     & 0.916     \\ 
\cline{2-7}
                                                                                & \multirow{3}{*}{AutoLoRA} & Top 1 & 17.521 & 0.329 & 6.300     & 0.920     \\
                                                                                &                           & Top 2 & 17.592 & 0.335 & 6.362     & 0.922     \\
                                                                                &                           & Top 3 & 17.644 & 0.340 & 6.401     & 0.922     \\ 
\midrule
\multirow{4}{*}{DiffusionDB}                                                    & \multicolumn{2}{c|}{FLUX.1-dev} & 17.887 & 0.315 & 6.425     & 0.849     \\ 
\cline{2-7}
                                                                                & \multirow{3}{*}{AutoLoRA} & Top 1 & 18.057 & 0.328 & 6.496     & 0.857     \\
                                                                                &                           & Top 2 & 18.109 & 0.332 & 6.522     & 0.861     \\
                                                                                &                           & Top 3 & 18.098 & 0.334 & 6.515     & 0.861     \\
\bottomrule
\end{tabular}
\end{table}

\begin{figure*}[t]
    \centering
    \includegraphics[width=0.9\linewidth]{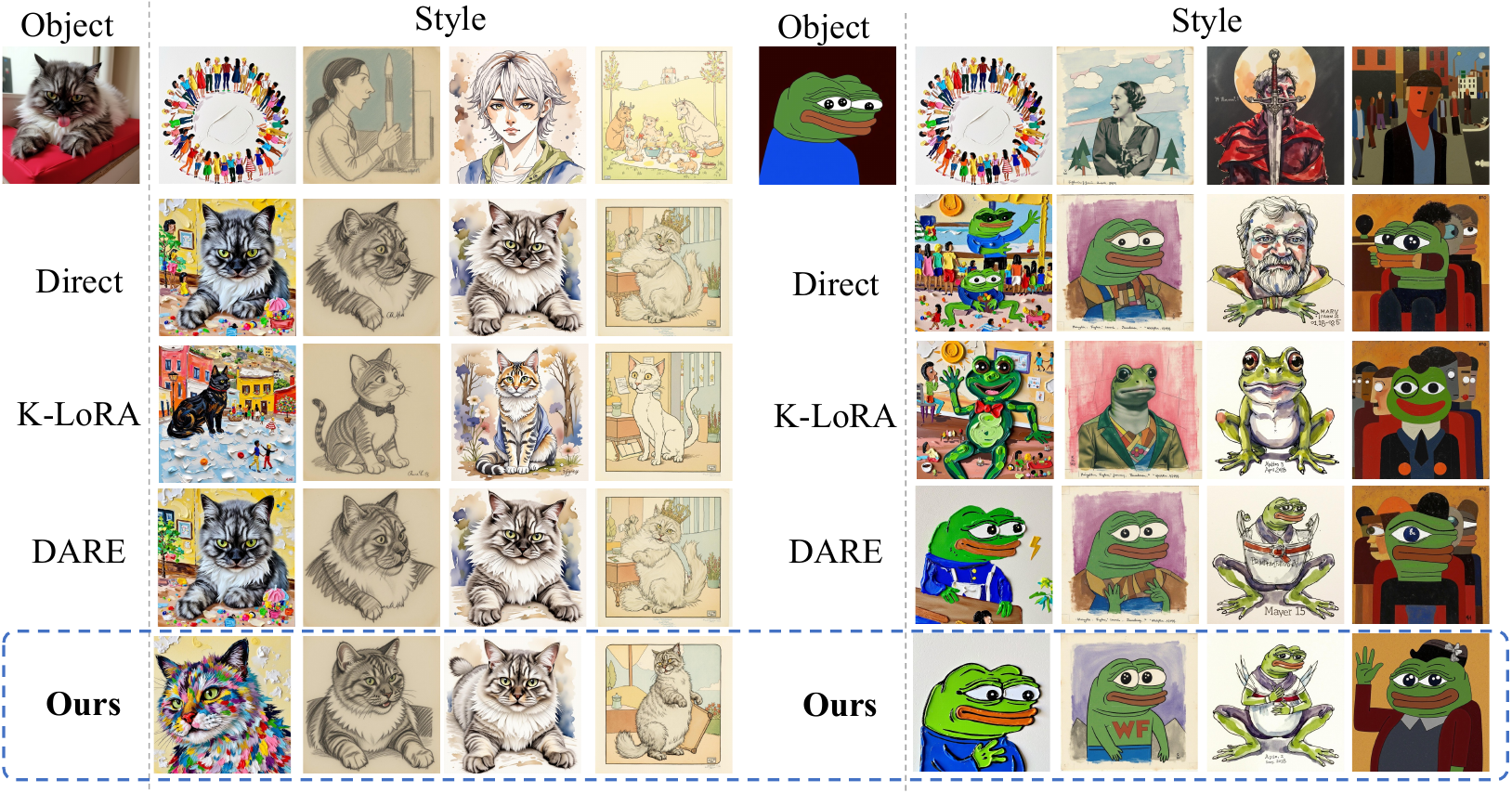}
    \caption{Qualitative comparison of object-style LoRA fusion. Compared with other baseline methods, our fine-grained gating fusion mechanism can seamlessly integrate multiple LoRAs while both ensuring object consistency and effectively preserving style attributes.}
    \label{fig:object-style-fusion}
\end{figure*}

\subsection{Effectiveness of Fine-Grained Gated Fusion}
In this section, we conduct a dedicated evaluation of the proposed Fine-Grained Gated Fusion mechanism's effectiveness through two complementary approaches: (1) fusing commonly used object and style LoRAs, and (2) randomly selecting 2–3 LoRAs from our candidate pool for fusion. 
\subsubsection{Experimental Setup} 
Following prior work K-LoRA\cite{Ziheng2025Klora}, we first evaluate fusion performance on object and style LoRAs separately, using the 3 object LoRAs and 8 style LoRAs provided by K-LoRA. For each object-style combination, we generate images using the template prompt:"a \{\textit{Object}\} in the \{\textit{Style}\} style", producing 10 images per prompt. Evaluation employs CLIP\cite{Alec2021clip} to compute similarity scores between generated images and both (1) object images and (2) style reference images.
Beyond fixed object-style fusion, we randomly select 2–3 LoRAs from our candidate pool and use Qwen VL to generate descriptive prompts based on their representative images, constructing a LoRA fusion test set. This scenario is evaluated using aesthetic scores and CLIP similarity metrics comparing generated images against those produced by individual LoRAs.

\subsubsection{Result}
From the results of Object and Style LoRA fusion in Table \ref{tab:obj-style-fusion}, it is evident that our fusion method significantly outperforms other baselines in simultaneously preserving object consistency and style similarity. Direct linear addition tends to prioritize stylistic attributes at the expense of object fidelity. Similarly, K-LoRA specifically amplifies the weights of Style LoRA during generation, causing style features to dominate.
Table \ref{tab:2=loras-fusion}  and Table \ref{tab:3-lora-fusion} present results from fusing randomly selected LoRAs. K-LoRA is designed exclusively for Object-Style LoRA fusion, leading to suboptimal performance in generalized scenarios. Moreover, its effectiveness depends on LoRA loading order—it assumes the second loaded LoRA is the style component and disproportionately emphasizes it. Both direct linear addition and DARE trigger mutual suppression between LoRAs, degrading image quality. This degradation becomes critical when fusing three LoRAs, where these methods completely fail to function.
In contrast, our approach remains robust when fusing any number of LoRAs. It maintains image quality while effectively integrating capabilities from LoRAs, demonstrating the efficacy of our fine-grained gated fusion mechanism. As show in Fig. \ref{tab:obj-style-fusion}, our method seamlessly integrates LoRAs. For instance, generating a cat rendered in oil painting style rather than a disjointed composition of a cat against an oil-painted background.

\begin{table}[ht]
\centering
\caption{Quantitative results of object and style LoRA fusion compared to baselines.}
\label{tab:obj-style-fusion}
\begin{tabular}{c|cc} 
\toprule
\textbf{Method} & \textbf{Obj Sim} ($\uparrow$) & \textbf{Style Sim} ($\uparrow$)  \\
\midrule
Direct          & 0.728            & 0.579               \\
K-LoRA          & 0.639            & 0.624               \\
DARE            & 0.732            & 0.577               \\
\textbf{Ours}   & 0.742            & 0.577               \\
\bottomrule
\end{tabular}
\end{table}

\begin{table}[ht]
\centering
\caption{Comparison of random two LoRA fusion. "Original" denotes image generation without LoRA, "$L_i$ Sim" represents similarity to the $i-th$ LoRA's generated image.}
\label{tab:2=loras-fusion}
\begin{tabular}{c|ccccc} 
\toprule
\textbf{Method}   & \textbf{MPS}   & \textbf{HPS}   & \textbf{VQA}   & \textbf{$L_1$ Sim} & \textbf{$L_2$ Sim}  \\
\midrule
Original & 18.28 & 0.330 & 0.917 & 0.822     & 0.822      \\
\midrule
Direct   & 17.19 & 0.305 & 0.891 & 0.854     & 0.840      \\
K-LoRA   & 17.40 & 0.311 & 0.897 & 0.845     & \textbf{0.887}      \\
DARE     & 16.41 & 0.285 & 0.864 & 0.832     & 0.818      \\
\textbf{Ours}     & \textbf{18.36} & \textbf{0.336} & \textbf{0.929} & \textbf{0.856}     & 0.848      \\
\bottomrule
\end{tabular}
\end{table}

\begin{table}
\centering
\caption{Comparison of random three LoRA fusion.}
\label{tab:3-lora-fusion}
\setlength{\tabcolsep}{3pt}
\begin{tabular}{c|cccccc} 
\toprule
\textbf{Method}   & \textbf{MPS}    & \textbf{HPS}   & \textbf{VQA}   & \textbf{$L_1$ Sim} & \textbf{$L_2$ Sim} & \textbf{$L_3$ Sim}  \\
\midrule
Original & 18.351 & 0.332 & 0.919 & 0.837  & 0.826  & 0.851   \\
\midrule
Direct   & 16.780 & 0.292 & 0.886 & 0.816  & 0.811  & 0.811   \\
DARE     & 14.880 & 0.249 & 0.812 & 0.773  & 0.768  & 0.811   \\
\textbf{Ours}     & \textbf{18.557} & \textbf{0.339} & \textbf{0.932} & \textbf{0.858}  & \textbf{0.851}  & \textbf{0.852}   \\
\bottomrule
\end{tabular}
\end{table}

\subsection{Effectiveness of LoRA Encoder}
\begin{figure}[ht]
    \centering
    \includegraphics[width=0.99\linewidth]{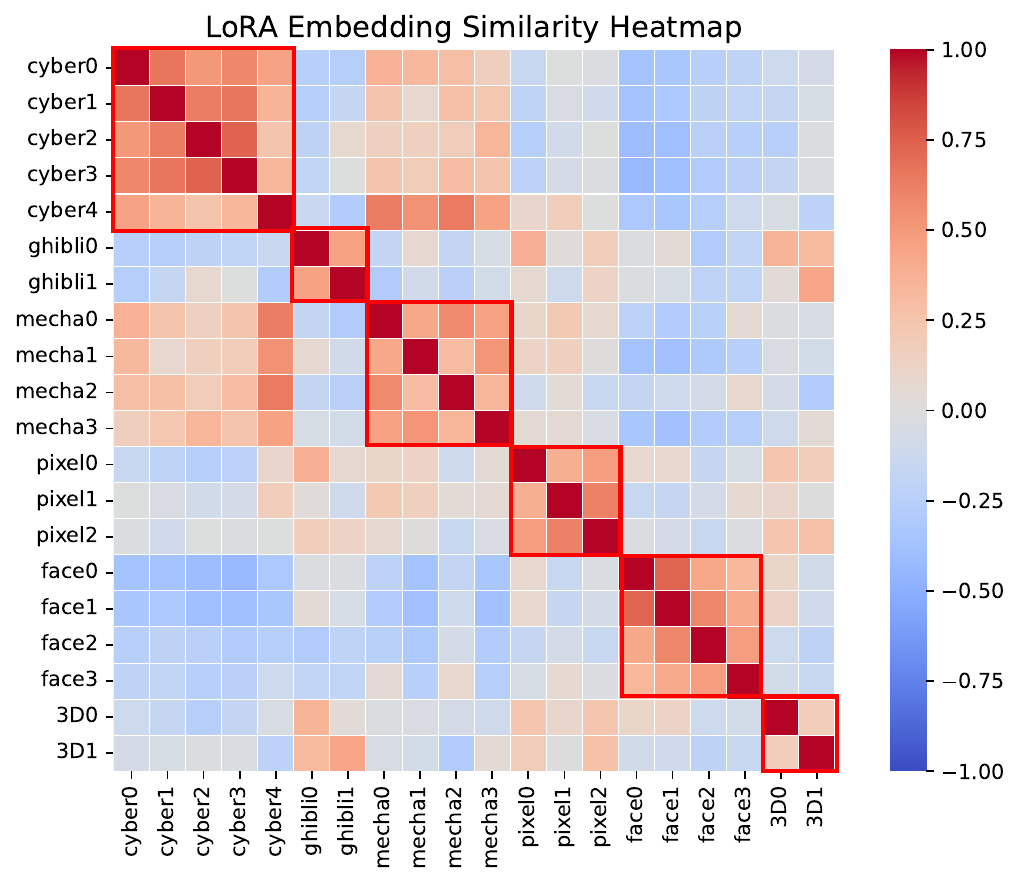}
    \caption{LoRA embedding similarity heatmap. LoRAs from the same theme are grouped in square brackets.}
    \label{fig:lora-sim}
\end{figure}

To validate the effectiveness of LoRA Encoder, we selected 21 LoRAs from the dataset spanning 6 distinct themes (including cyberpunk style, Ghibli Style, Mecha, Pixel Style, Enhance Face, and 3D Rendering) and computed pairwise embedding similarities across all combinations. As visualized in Fig. \ref{fig:lora-sim} through a heatmap, LoRAs belonging to the same theme are grouped within identical bounding boxes. This visualization clearly demonstrates that LoRAs of identical themes exhibit significantly higher embedding similarity, confirming that our LoRA Encoder successfully encodes the semantically related LoRAs into adjacent regions of the feature space. Consequently, these embeddings enable highly precise LoRA retrieval in practical applications.

\subsection{Ablation Studies}
\subsubsection{Impact of global LoRA}
This study investigate the impact of using a constructed Global LoRA for training and inference on LoRA fusion performance. We retrain the gated fusion model on 162 LoRA models without Global LoRA integration while maintaining all other configurations. We conduct experiments on three LoRA fusion datasets, with the results shown in Table \ref{tab:global-lora}. Experimental results demonstrate that adding Global LoRA to training and generation significantly improves the overall performance of multi-LoRA fusion, particularly in the aesthetic scores of generated images. This demonstrates that Global LoRA helps capture and integrate global information across multiple LoRAs during fusion, reducing conflicts. We also observed improvements in individual LoRA similarity, suggesting that Global LoRA allows the fusion module to better focus on the unique aspects of each LoRA. 



\begin{table}
\small
\centering
\caption{ The impact of Global LoRA integration on LoRA fusion performance was evaluated on the tri-LoRA dataset.  }
\label{tab:global-lora}
\setlength{\tabcolsep}{3pt}
\begin{tabular}{c|cccccc} 
\toprule
\textbf{Method}                                                    & \textbf{MPS}    & \textbf{HPS}   & \textbf{VQA} & \textbf{$L_1$ Sim} & \textbf{$L_2$ Sim} & \textbf{$L_3$ Sim}  \\
\midrule
\begin{tabular}[c]{@{}c@{}}W.o/\\Global LoRA\end{tabular} & 17.869 & 0.324 & 0.916    & 0.849     & 0.839     & 0.844      \\
\begin{tabular}[c]{@{}c@{}}W/\\Global LoRA\end{tabular}   & \textbf{18.557} & \textbf{0.339} & \textbf{0.932}    & \textbf{0.858}     & \textbf{0.851}     & \textbf{0.852}      \\
\bottomrule
\end{tabular}
\end{table}

\section{Conclusion}
In this study, we investigate how to leverage the vast repository of community-sourced LoRA models to enhance the performance of text-to-image generation model. The core challenge lies in the dynamically evolving LoRA pool, where open-source LoRAs frequently suffer from inadequate documentation, and significant hurdles in efficiently fusing multiple LoRAs. To address these limitations, we propose AutoLoRA, a novel framework that first employs a Weight Encoding-based LoRA Retriever to retrieve relevant models through text-LoRA embedding similarity computation. Subsequently, it integrates a Fine-Grained Dynamic Gated Fusion mechanism that dynamically calculates LoRA weights across different dimensions via a learnable gated module, thereby optimizing the collaborative performance of multiple LoRAs. Empirical results validate the effectiveness of AutoLoRA, demonstrating its potential to facilitate the development and utilization of the rapidly expanding ecosystem of open-source LoRA models.

\bibliography{aaai2026}

\end{document}